\documentclass[letterpaper, 10 pt, journal, twoside]{IEEEtran}

\IEEEoverridecommandlockouts                              

\usepackage{graphics} 
\usepackage{epsfig} 

\usepackage[table,dvipsnames]{xcolor}
\usepackage{booktabs}
\usepackage{bbm}
\usepackage{graphicx}
\usepackage{longtable}
\usepackage{flushend}
\usepackage{subcaption}
\usepackage{bm,stmaryrd}
\usepackage{soul}
\newcommand\shortminus{\mathop{\mbox{-}}}

\makeatletter
\let\NAT@parse\undefined
\makeatother
\usepackage[pagebackref=false,colorlinks=true,citecolor=blue,linkcolor=blue]{hyperref} 

\usepackage{amssymb}
\usepackage{pifont}%
\newcommand{\bx}{\bm{x}}

\newcommand{\cL}{\mathcal{L}}

\newcommand{\Tu}{\mathcal{D}^{\mathrm{u}}}  
\newcommand{\Tl}{\mathcal{D}^{\ell}}        
\newcommand{\Tp}{\mathcal{D}^{\mathrm{p}}}  

\newcommand{\teach}{T}
\newcommand{\teachn}{\teach_{\shortminus n: n}}
\newcommand{\teachon}{\teach_{\shortminus n: 0}}

\newcommand{\stud}{S}

\newcommand{\concordance}{\mathcal{T}}

\usepackage{amssymb}
\usepackage{amsmath}
\DeclareMathOperator*{\argmax}{argmax}

\usepackage{dsfont}

\definecolor{person-color}{RGB}{255,30,30}
\definecolor{road-color}{RGB}{255,0,255}
\definecolor{side-walks-color}{RGB}{75, 0, 75}
\definecolor{bicycle-color}{RGB}{100, 230, 245}
\usepackage{xcolor}
\newcommand\crule[3][black]{\textcolor{#1}{\rule{#2}{#3}}}

\DeclareUnicodeCharacter{0301}{\'{e}}

\title{\LARGE \bf
Teachers in concordance for pseudo-labeling of 3D sequential data
}

\author{Awet Haileslassie Gebrehiwot\thanks{*~Corresponding author: E-mail: awethaileslassie21@gmail.com}$^{1*}$, Patrik Vacek$^{1}$, David Hurych$^{2}$, Karel Zimmermann$^{1}$, Patrick Perez$^{2}$, \\Tomáš Svoboda$^{1}$ \\%
\emph{$^{1}$~Czech Technical University in Prague, FEE, Dept of Cybernetics}\\%
\emph{$^{2}$~Valeo.ai}%

\thanks{
\emph{}}%

}

\makeatletter
\def\ps@IEEEtitlepagestyle{%
  \def\@oddfoot{\mycopyrightnotice}%
  \def\@oddhead{\hbox{}\@IEEEheaderstyle\leftmark\hfil\thepage}\relax
  \def\@evenhead{\@IEEEheaderstyle\thepage\hfil\leftmark\hbox{}}\relax
  \def\@evenfoot{}%
}
\def\mycopyrightnotice{%
  \begin{minipage}{\textwidth}
  \centering \scriptsize
  This work has been submitted to the IEEE for possible publication.
  \end{minipage}
}
\makeatother

\begin{document}

\maketitle

\begin{abstract}
Automatic pseudo-labeling is a powerful tool to tap into large amounts of sequential unlabeled data. It is especially appealing in safety-critical applications of autonomous driving, where performance requirements are extreme, datasets are large, and manual labeling is very challenging. We propose to leverage sequences of point clouds to boost the pseudo-labeling technique in a teacher-student setup via training multiple teachers, each with access to different temporal information. This set of teachers, dubbed \emph{Concordance}, provides higher quality pseudo-labels for student training than standard methods. The output of multiple teachers is combined via a novel pseudo-label confidence-guided criterion. Our experimental evaluation focuses on the 3D point cloud domain and urban driving scenarios. We show the performance of our method applied to 3D semantic segmentation and 3D object detection on three benchmark datasets. Our approach, which uses only 20\% manual labels, outperforms some fully supervised methods. A notable performance boost is achieved for classes rarely appearing in training data. 
Our codes are publicly available on \url{https://github.com/ctu-vras/T-Concord3D}.
\end{abstract}


\section{Introduction}
\label{sec:intro}
In many machine learning problems, state-of-the-art performance requires supervision via complete annotation of the training data.
Therefore, an effective way to increase the performance of a model is to add more annotated training data \cite{jiang2021guided}. However, this approach is neither scalable nor sustainable, requiring thorough manual labeling by human annotators. 
Annotation tasks such as semantic segmentation of videos and sequences of point clouds are very complex to accomplish.
Reducing annotation needs is, therefore, a crucial and active research field. 
\textit{Pseudo-labeling} \cite{Lee2013PseudoLabelT} has emerged as a versatile and powerful tool. A teacher model trained on a small amount of labeled data is used to annotate lots of unlabeled data automatically. The student model trains on the combination of a small labeled set and a large pseudo-labeled set.
This work introduces ways to boost pseudo-labeling when dealing with temporally ordered data streams. The proposed framework is instantiated and evaluated in the context of 3D point-cloud analysis for driving applications.
\begin{figure}[t]
    \centering
    \includegraphics[width=0.99\linewidth,keepaspectratio]{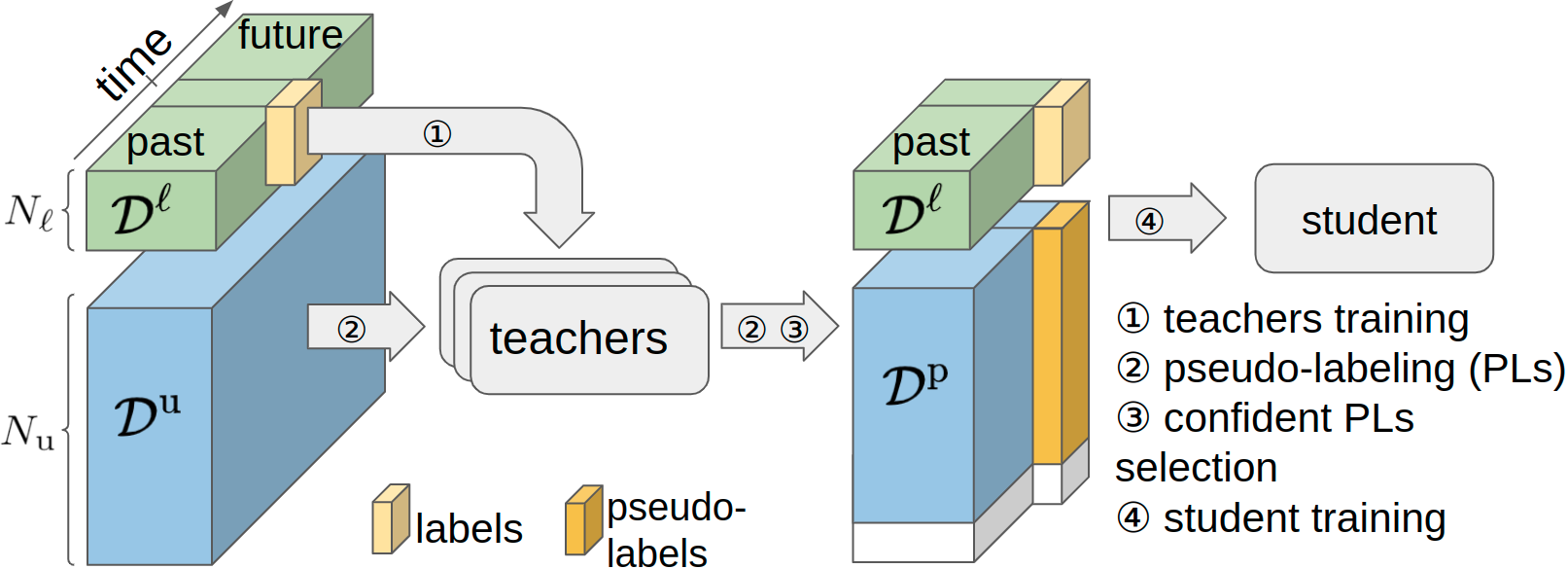}
    \caption{\textbf{Proposed ``Concordance of teachers''  for pseudo-labeling of sequences}.
    A set $\Tl$ of sequences with a central frame labeled and a larger set $\Tu$ of unannotated ones are available for training; \raisebox{.9pt}{\textcircled{\raisebox{-.9pt}{1}}} Multiple offline teachers are trained with full supervision on $\Tl$, each with a different temporal range towards future and past frames; \raisebox{.9pt}{\textcircled{\raisebox{-.9pt}{2}}} The teachers are run on $\Tu$ to produce pseudo-labels (PLs) for central frames; \raisebox{.9pt}{\textcircled{\raisebox{-.9pt}{3}}} Sequences with the most confident PLs according to Concordance of teachers are selected, forming the pseudo-labeled set $\Tp$. The white box depicts the discarded PLs; \raisebox{.9pt}{\textcircled{\raisebox{-.9pt}{4}}} The student is trained on $\Tl\cup\Tp$, to work online with past and current frames only.}
    \label{fig:pipeline}
\end{figure}
In contrast to unordered data, the temporal ordering of the data grants the student and teacher access to a meaningful temporal context. In particular, the teacher can also access future data in the form of privileged information \cite{VAPNIK2009544} that is available at the time of pseudo-labeling but not at the students' inference time.
Consequently, the teacher can benefit from past and future frames, thus making the most of the temporal consistency over an extended time window. 
We noticed that the range of this temporal window has a crucial influence on teachers' performance, as it captures different temporal contexts. 
The complexity of spatio-temporal events in 3D driving scenes would require the teacher to operate simultaneously at different temporal ranges. Learning a large enough teacher capable of modeling the aforementioned complexity would require many labels, contradicting the motivation of easing the annotation. We take a more practical approach where multiple complementary teachers are trained, each operating in its own temporal range with the past and future frames unannotated. 
These teachers use the \emph{Concordance} to assess the confidence of the extracted pseudo-labels (PLs) and to select the most confident ones for student training eventually.

We experimentally demonstrate that Concordance-based pseudo-labeling (i) achieves competitive performance with state-of-the-art fully supervised methods \cite{xu2020squeezesegv3,Milioto2019RangeNetF,zhang2020polarnet,zhou2020cylinder3d} with only a fraction of labeled data, (ii) outperforms pseudo-labeling methods that do not leverage multiple teachers \cite{jiang2021guided}.

Our approach (Fig.\,\ref{fig:pipeline}) only assumes that two sets of sequences are available: The first with the central frames annotated and the second larger and devoid of annotation. Several teachers are trained on the first set to predict the label of the middle frame of an input sequence. Once trained, all the offline models run on the second dataset to pseudo-label the central frames. The most promising automatically annotated samples are weighted and added to the first set based on time-aware Concordance sample selection. The resulting large labeled set, with the future frames not available at the input, is used to train the final online model.    

\begin{figure*}[t]
    \centering
    \includegraphics[width=0.8\linewidth,keepaspectratio]{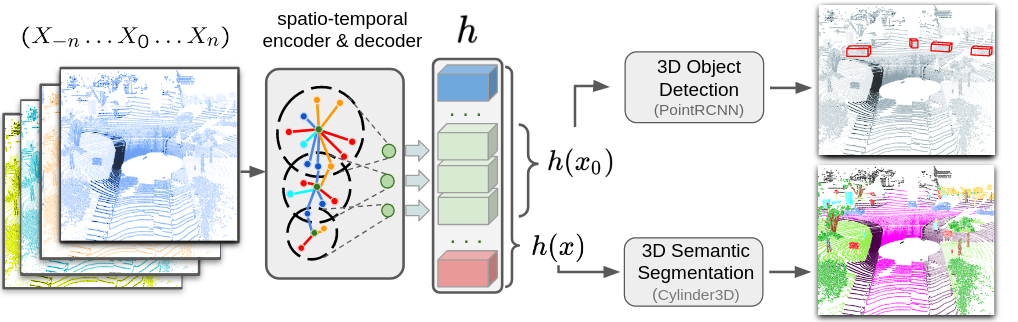}
    \caption{\textbf{Proposed architectures that aggregate sequence of 3D point clouds}. We build a modified version of Cylinder3D for semantic segmentation and PointRCNN for object detection that can aggregate multiple frames inside its spatio-temporal encoder. The output of the spatio-temporal encoder-decoder are extracted point features $\bm{h}(\bx)$. For object detection, Only features from the reference frame, i.e., $\bm{h}(\bm{x_0})$ for $\bx_0\in X_0$, are used to train PointRCNN for object detection. Here, we use the full set of features $\bm{h}(\bx)$ for semantic segmentation due to the setup of the state-of-the-art Cylinder3D architecture.
    }
    \label{fig:pointrcnn}
\end{figure*}

We put this framework to work for different spatio-temporal perception tasks on sequences of outdoor point clouds (PCs). We take advantage of the temporal ordering 
to provide more accurate pseudo-labels than an ordinary PL method would deliver. We demonstrate its superiority on the tasks of 3D detection and 3D semantic segmentation in driving scenes on two architectures~\cite{shi2019pointrcnn,zhou2020cylinder3d} and three datasets (see Fig.\,\ref{fig:pointrcnn}). 
Our contributions to pseudo-labeling of temporal data are:
1) An effective way to aggregate time-ordered unannotated/annotated 3D scans; Leveraging such privileged information improves teacher's performance for 3D semantic segmentation and object detection tasks.
2) A novel confidence-guided criterion for better pseudo-labels selection and loss function guidance.
3) A novel weighting of pseudo-labels via the Concordance of teachers trained on different temporal ranges.

\section{Related Work}
\label{sec:related}
\smallskip\noindent\textbf{Spatial and temporal consistency in point clouds.} LiDAR PCs are unstructured data. PointNet architecture~\cite{journals/corr/pointnet} directly consumes raw PCs to extract features with permutation invariance and global features for classification. PointNet++~\cite{journals/symmetry/pointnet++} further improves the extraction of local features. The architecture of MeteorNet~\cite{Meteornet} works with multiple input PCs to extract features from additionally available points with consistent temporal properties. Choy \textit{et al}. \cite{Minknet} use sparse 4D CNN~ for spatiotemporal perception to improve robustness in detection tasks. Spatial synchronization to the reference scan and adding one additional channel of encoded time lead to further performance gain~\cite{what_you_see}. Qi \textit{et al}.~\cite{off-board_waymo} use temporal information for PC densification based on tracking previously detected objects for automatic data annotation. Conversely to us, they use a top-performance detection model pre-trained in a fully supervised way. We focus on building a set of teacher models (Concordance) from a minimal amount of annotated data and apply distillation through pseudo-labeling. 

\smallskip\noindent\textbf{Knowledge distillation.} Training the student model is usually done by distilling knowledge in the feature or output space \cite{Knowledge_orig}. Liu \textit{et al}.~\cite{Knowledge_flow} distill an ensemble of teachers into a single student and extend the idea using different architectures suitable for different tasks as teachers. Cho and Hariharan~\cite{Knowledge_efficiancy_2019_ICCV} show that larger models are not necessarily better teachers, mainly due to parameter complexity mismatch. Mirzadeh \textit{et al}.~\cite{Teacher_assistant} propose a multistep knowledge distillation with an intermediate-sized network to bridge the gap between student and teacher complexity. The benefit of using the teacher network can also come from exploiting privileged information~\cite{automatic-labeling2022}.
In our work, we adopt a teacher-student framework and distill future data in sequential frames when training the teacher models.

\smallskip\noindent\textbf{Semi-supervised learning.} 
Semi-supervised learning approaches have been heavily researched for image recognition, less so for point clouds~\cite{jiang2021guided, SESS_2020_CVPR}. Extending an image pseudo-labeling approach~\cite{Lee2013PseudoLabelT} to 3D perception~\cite{automatic-labeling2022} has recently shown that automatic labeling of LiDAR data could be leveraged to achieve considerable performance gain.
An early approach~\cite{tracking-based2012} proposes to extract useful training examples from unlabeled data by exploiting the temporal information in LiDAR scans for classification.
Enforcing consistency of model predictions across perturbed versions of unlabelled data~\cite{SESS_2020_CVPR} proves to be beneficial in 3D object detection. For the task of 3D semantic segmentation, learning with a point-guided contrastive loss~\cite{jiang2021guided} increases performance even with fewer ground-truth labels. The authors show that using pseudo-labels and confidence thresholding can help to improve feature learning. This method is compared to ours in Section~\ref{sec:exp}.
Our approach focuses on semi-supervised learning using pseudo-labeling and knowledge distillation. It is worth mentioning that a complementary line of work explores approaches such as self-supervised pre-training~\cite{xie2020pointcontrast,SegContrast2022} and domain adaption~\cite{jaritz2020xmuda,LiDARNet} to avoid labeling too many samples.

\section{Method}
\label{sec:method}

\smallskip\noindent\textbf{Point-cloud notations.} 
A point cloud $X = \{\bx^{k}\}_{k=1}^{K}$
is a finite order-less collection of 3D points, where the number of points $K$ is assumed constant over time to keep the notation simple. We consider symmetric time-ordered PC sequences of the form $X_{\shortminus n:n} = (X_{\shortminus n}, \ldots, X_{\shortminus 1}, X_{0}, X_{1}, \ldots, X_{n})$ 
composed of a \textit{reference scan} $X_0$, preceded by $n$ past scans and followed by $n$ future ones. For the symmetric sequences, the reference scan (frame) is the same as the central one, see Fig.\,\ref{fig:pipeline}. All scans in the sequence are transformed into the coordinate system of the reference one.

\smallskip\noindent\textbf{Time-aware feature extraction.}
\label{pointnet++dirct} 
We build on a modified architectures of the Cylinder3D \cite{zhou2020cylinder3d} for semantic segmentation and PointRCNN \cite{shi2019pointrcnn} for object detection. Both architectures consist of MLP modules responsible for attaching rich spatial features to individual scan points, followed by task-dependent modules. In contrast to single-frame perception, we must handle successive scans; therefore, we propose a time-aware extension of their original MLPs.

\begin{figure}[t!]
    \centering
    \includegraphics[width=\linewidth,keepaspectratio]{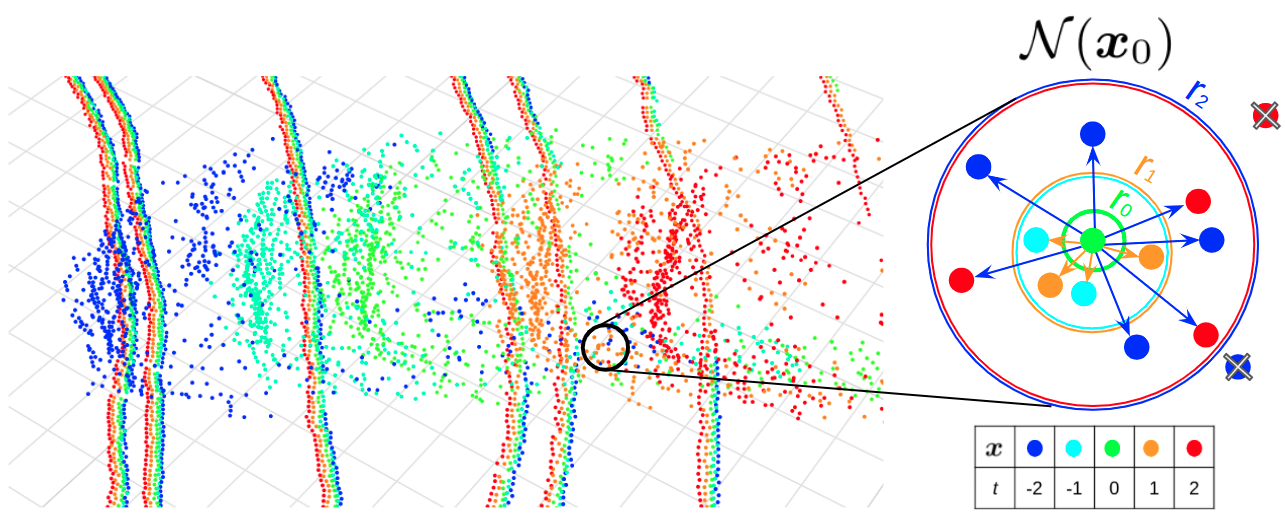}
    \caption{\textbf{Time-aware neighborhood of a point in the reference scan}. Points from different times are presented in different colors, where circled green point ${\bx}_0$ is from the (green) reference frame $X_0$. Its spatio-temporal neighborhood $\mathcal{N}({\bx}_0)$ is composed of all points in scan $X_t$ in a spatial radius $r(|t|)$, for $t=-n, \cdots , n$. Crossed points are excluded from this spatio-temporal neighborhood.}
    \label{fig:temporal-feature-learning}
\end{figure}

Given a sequence $X_{\shortminus n:n}$ of point clouds, the backbone architecture estimates a feature vector $\bm{h}(\bx_0)$ for each point $\bx_0$ in the reference scan $X_0$. This feature vector encompasses all contextual information from its spatio-temporal neighborhood $\mathcal{N}(\bx_0)$ defined as an hourglass-like 3D shape centered in $\bx_0$: 
\begin{equation}
    \mathcal{N}(\bx_0)\!=\!\big\{\bx_t \!\in \!X_t: \|\bx_t\!-\!\bx_0\|\leq r(|t|), \; t\in\llbracket-n,n\rrbracket \big\},
    \label{eq:neighborhood}
\end{equation}
where $r$ is an increasing function as in \cite{Meteornet} (Fig.\,\ref{fig:temporal-feature-learning}). The feature of each $\bx_0 \in X_{0}$ is finally defined as:
\begin{equation}
\bm{h}(\bx_0) = \max_{\bx_t\in \mathcal{N}(\bx_0)}
\big\{\phi(\bx_t-\bx_0,t)\big\},
\end{equation}
that is, by max-pooling of time-aware pairwise features over the spatio-temporal neighborhood, where $\phi$ is an MLP with shared weights to be trained, and $t \in\llbracket -n, n \rrbracket$.

The construction of the spatio-temporal neighborhoods obeys the intuition that the maximum distance an object can travel between two scans increases with the object's speed and the temporal separation between the scans. Thus, the maximum spatial distance for grouping points should increase with their temporal distance. 

\smallskip\noindent\textbf{Task-dependent modules.} 
The feature extraction method provides point-wise feature vectors and their corresponding 3D positions for the points in the reference coordinate frame of $X_0$. The feature vectors are then fed into subsequent task-dependent modules.

\textit{3D Object Detection}: We consider the 3D detection of vehicles with our multi-frame adaptation of PointRCNN~\cite{shi2019pointrcnn}. Here, the bounding box labels are not associated with specific input points but with a specific 3D position. If we consider box labels from all frames, we would not know which points on input are associated with each one. Therefore, we use the bounding box labels only for the reference scan $X_{0}$ and mask the box labels from other frames. This differs from semantic segmentation, where each point has its own exclusive class probability in each frame.

\textit{3D Semantic Segmentation}: We adopt Cylinder3D \cite{zhou2020cylinder3d} and extend it into a semi-supervised approach. A trained Cylinder3D classifier semantically labels each point. Here, we use labels for all temporal input instants.

\smallskip\noindent\textbf{Training data.} 
We consider two types of teacher models for training: (i) $\teachn$ with access to future frames, and (ii) $\teachon$, which has access only to past frames.

\begin{figure}[t]
    \centering
    \includegraphics[width=0.43\textwidth,height=0.27\textheight]{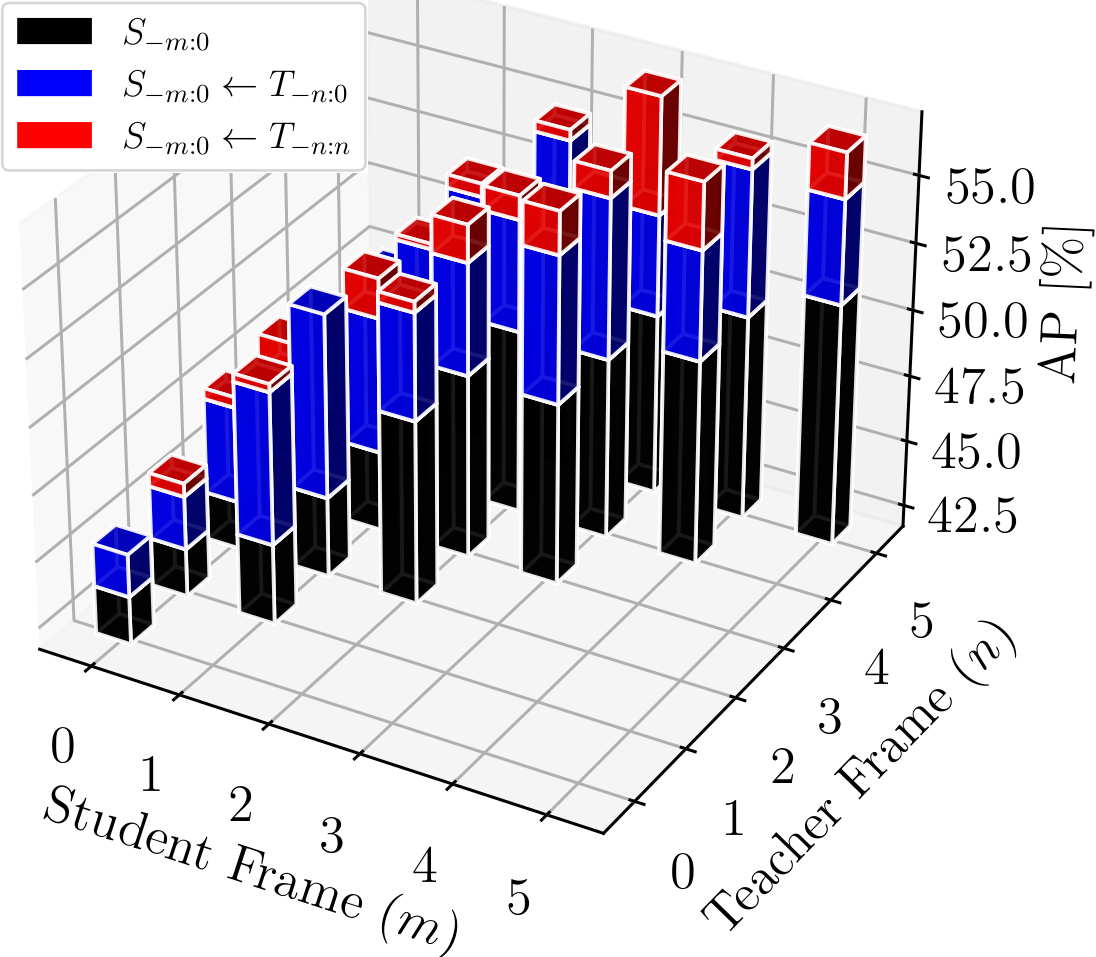}
    \caption{\textbf{Impact of knowledge distillation}. Performance of distilling privileged information from a single teacher into a student in the 3D object detection task.}
    \label{fig:all-student-teacher}
\end{figure}

To understand the effect of distilling a teacher model with access to privileged information $\teach_{\shortminus n:n}$, we have performed an experiment where we train a student model by the pseudo-labels provided by one teacher with and without privileged information. As shown in Fig.\,\ref{fig:all-student-teacher} distilling a single teacher with privileged information in a student model $\stud_{\shortminus m:0} \shortleftarrow \teach_{\shortminus n:n}$ provides the best performance (red on top) over the baseline supervised student (black pillars) and over distilling a single teacher without privileged information into a student model $\stud_{\shortminus m:0} \shortleftarrow \teach_{\shortminus n:0}$ (blue pillars).

\smallskip\noindent\textbf{Concordance and selection of pseudo-labels.} Inspired by our finding from Fig.\,\ref{fig:all-student-teacher}, we want to fully exploit the information contained in the available scan sequences of length $(2n+1)$. We propose Concordance of teachers, where we train a set of $n$ teachers with a varying span of the temporal context. It means that teacher $T_{\shortminus 1:1}$ is trained on the subsequences $X_{\shortminus 1:1}$, teacher $T_{\shortminus 2:2}$ on $X_{\shortminus 2:2}$ and so on, for the sake of simplicity, we further drop the distinction between these sequences and assume that given a sequence $X\in\mathcal{D}^\textrm{u}$ any network will crop the temporal range appropriately.
Given the Concordance of teachers 
\begin{equation}
    \mathcal{T}^{\{1, \dots, n\}} = \{\teach_{\shortminus 1, 1}, \teach_{\shortminus 2, 2}, \dots, \teach_{\shortminus n, n} \}.
\label{eq:concordance}    
\end{equation}
%
%
We independently process every unlabeled sequence $X\in\mathcal{D}^\textrm{u}$ by all trained teachers.
Since the nature of outputs is slightly different for the semantic segmentation (point-wise class probabilities) and for the object detection (3D bounding boxes with class probabilities), we split the Concordance description at this point to avoid any misinterpretation.

\noindent\textbf{Concordance for semantic segmentation.} We estimate the pseudo-label $k^*$ and its confidence $c$ for each output point. Given the point, each teacher $T\in \mathcal{T}^{\{1, \dots, n\}}$ provides a vector of class probabilities ${\hat{\mathbf{y}}^{\teach}}$. 
We then estimate the teacher-wise pseudo-labels $k^{*}(\teach)$
\begin{equation}
     k^{*}(\teach) = \argmax_{k \in K} {\hat{\mathbf{y}}^{\teach}}_{k}.
     \label{eq:pseudo-label}
\end{equation}
We pay special attention to the teacher with the strongest opinion, the one with the highest output value $\hat{\mathbf{y}}_k^T$. In particular, we denote this teacher as $T^*$, its pseudo-label $k^*$ and the score of this pseudo-label as $y^*= \hat{\mathbf{y}}_{k^*}^{T^*}$.

We determine the pseudo-label of the given output point as $k^*$.
The confidence of $k^*$ is defined as the weighted sum of two criteria: (i) the score $y^*$ of the pseudo-label and (ii) the number of other teachers $\teach$ that predict the same class, \emph{i.e.}, for which $k^{*} = k^{*}(\teach)$.
The trade-off between these two criteria is determined by a non-negative weight $\lambda$ as follows: 
\begin{equation}
     \hat{c} = y^* + \lambda  \sum_{\teach \in \concordance\setminus T^*} \mathds{1} \llbracket k^{*} = k^{*}(\teach) \rrbracket\,.
     \label{eq:computed-weight}
\end{equation}
To preserve compatibility with training loss Eq.~(\ref{eq:mean-loss}), this confidence is clipped: 
    $c  = \min (1, \hat{c})\,.$

\noindent\textbf{Concordance for object detection.} Each teacher provides a different set of 3D bounding boxes (bbs) and class probabilities. To calculate final pseudo-label confidence,
we need to extract bbs that 
correspond to a single physical object. We greedily search for clusters of bbs with mutual intersection-over-union above a user-defined threshold. The algorithm starts by building the first cluster from the strongest bb. Once there are no more bbs with a sufficient IoU with the strongest bb, we stop building the cluster, suppress all associated bbs, and continue building a following cluster from the remaining bbs. The class probabilities $\mathbf{y}^T$ corresponding to every single cluster are then used directly to compute the pseudo-label and its confidence by the same procedure as for the semantic segmentation.
Following the standard practice in pseudo-labeling (\textit{e.g.}, \cite{DBLP:pseudo_objdet} in object detection or \cite{Li2020SelfLoopUA} in semantic segmentation), the final selection of pseudo-labels is obtained by thresholding the confidence. Individual pseudo-labels with confidence below the chosen threshold are masked out of the loss function and treated as a \textit{Don't Care} class. The final selection of pseudo-labels and associated data form the training set $\Tp$. 

\smallskip\noindent\textbf{Training the student.} The student model only performs inference online and, therefore, is learned only with past input sequences. Following pseudo-labeling through the teacher-student framework \cite{Lee2013PseudoLabelT,Shi_2018_ECCV}, we train the student model on both the human-labeled set and the pseudo-labeled one. The loss function for training the student is summed as follows:
\begin{equation}
    \cL = \frac{1}{M} \sum_{(\mathbf{y}, c) \in \Tp \cup \Tl} c ~ \cL_{\text{task}}(\hat{\mathbf{y}}, \mathbf{y}),
    \label{eq:mean-loss}
\end{equation}
where $\mathbf{y}$ is one hot label encoding vector, $\mathbf{\hat{y}}$ are model predictions, and $c$ is the corresponding confidence from our Concordance selection. Samples collected from the original human-labeled dataset have $c=1$. The original loss function of the task-dependent module is denoted $\cL_{\text{task}}$. All data are sampled from the combined datasets $\Tp$ and $\Tl$, and $M$ is the number of samples in the union.

 \section{Experiments}
 \label{sec:exp}
%
%
\begin{table*}[t]
\centering%
\caption{\textbf{LiDAR semantic segmentation performance on SemanticKITTI \textit{single-scan} test set.} Our method, $\stud_{0:0} \!\shortleftarrow\! \concordance^{\{1,2,3\}}$ (`Ours (20\%)') utilizes only 20\% of the labeled data, the remaining 80\% of training data being automatically annotated, to achieve a performance comparable with state-of-the-art methods. `Cylinder3D (20\%)' denotes Cylinder3D~\cite{zhou2020cylinder3d} trained, like our method, with only 20\% of labeled data; all other results are obtained from the literature, where full (100\%) labeled data is used. Performance in IoU percentages, per class and averaged, the higher, the better. Green and red indicate fully-supervised methods that have performance below and above the performance of the proposed method, respectively. `*' means that techniques such as fine-tuning and test-time augmentation (TTA) with flip and rotation are applied.} 
\setlength{\tabcolsep}{4.pt}{
\begin{tabular}{r@{\hskip 1em}r@{\hskip 0.8em}ccccccccccccccccccc}%
    \toprule%
    & \rotatebox[origin=l]{90}{{\textbf{mIoU}}} & {\rotatebox[origin=l]{90}{car}} & {\rotatebox[origin=l]{90}{bicycle}} & {\rotatebox[origin=l]{90}{motorcycle}} & {\rotatebox[origin=l]{90}{truck}} & {\rotatebox[origin=l]{90}{o-vehicle}} & {\rotatebox[origin=l]{90}{person}} & {\rotatebox[origin=l]{90}{bicyclist}} & {\rotatebox[origin=l]{90}{motorcyclist}} & {\rotatebox[origin=l]{90}{road}} & {\rotatebox[origin=l]{90}{parking}} & {\rotatebox[origin=l]{90}{sidewalk}} & {\rotatebox[origin=l]{90}{o-ground}} & {\rotatebox[origin=l]{90}{building}} & {\rotatebox[origin=l]{90}{fence}} & {\rotatebox[origin=l]{90}{vegetation}} & {\rotatebox[origin=l]{90}{trunk}} & {\rotatebox[origin=l]{90}{terrain}} & {\rotatebox[origin=l]{90}{pole}} & {\rotatebox[origin=l]{90}{traffic-sign}} \\
    \midrule%
    RangeNet++~\cite{Milioto2019RangeNetF}   & \textcolor{Green}{52.2}  & 91.4  & 25.7  & 34.4  & 25.7  & 23.0  & 38.3  & 38.8  & 4.8  & 91.8  & 65.0  & 75.2  & 27.8  & 87.4  & 58.6  & 80.5  & 55.1  & 64.6  & 47.9  & 55.9   \\
    PolarNet~\cite{zhang2020polarnet}     & \textcolor{Green}{54.3}  & 93.8  & 40.3  & 30.1  & 22.9  & 28.5  & 43.2  & 40.2  & 5.6  & 90.8  & 61.7  & 74.4  & 21.7  & 90.0  & 61.3  & 84.0  & 65.5  & 67.8  & 51.8  & 57.5  \\
    SqueezeSegV3~\cite{xu2020squeezesegv3}   & \textcolor{Green}{55.9}  & 92.5  & 38.7  & 36.5  & 29.6  & 33.0  & 45.6  & 46.2  & 20.1  & 91.7  & 63.4  & 74.8  & 26.4  & 89.0  & 59.4  & 82.0  & 58.7  & 65.4  & 49.6  & 58.9   \\
    KPConv~\cite{thomas2019kpconv}   & \textcolor{Green}{58.8}  & 96.0  & 32.0  & 42.5  & 33.4  & 44.3  & 61.5  & 61.6  & 11.8  & 88.8  & 61.3  & 72.7  & 31.6  & 95.0  & 64.2  & 84.8  & 69.2  & 69.1  & 56.4  & 47.4  \\
    Cylinder3D~\cite{zhou2020cylinder3d}   & \textcolor{red}{61.8}  & 96.1  & 54.2  & 47.6  & 38.6  & 45.0  & 65.1  & 63.5  & 13.6  & 91.2  & 62.2  & 75.2  & 18.7  & 89.6  & 61.6  & 85.4  & 69.7  & 69.3  & 62.6  & 64.7   \\
    (AF)2-S3Net~\cite{(AF)2-S3Net}* & \textcolor{red}{69.7} & 94.5 & 65.4 & 86.8 & 39.2 & 41.1 & 80.7 & 80.4 & 74.3 & 91.3 & 68.8 & 72.5 & 53.5 & 87.9 & 63.2 & 70.2 & 68.5 & 53.7 & 61.5 & 71.0 \\
    PVD~\cite{point-to-Voxel}* & \textcolor{red}{71.2} & 97.0 & 67.9 & 69.3 & 53.5 & 60.2 & 75.1 & 73.5 & 50.5 & 91.8 & 70.9 & 77.5 & 41.0 & 92.4 & 69.4 & 86.5 & 73.8 & 71.9 & 64.9 & 65.8 \\
    \midrule%
    Cylinder3D (20\%) & 51.9 & 92.1 & 31.7 & 28.5 & 25.1 & 22.1 & 49.6 & 32.4 & 26.4 & 86.9 & 47.2 & 67.5 & 12.6 & 88.7 & 55.7 & 83.2 & 64.4 & 64.8 & 53.4 & 53.9 \\
    \rowcolor{gray!30}Ours (20\%) & 58.9 & 92.9 & 46.4 & 36.6 & 35.1 & 27.3 & 62.4& 54.0 & 24.0 & 90.0 & 60.8 & 72.1 & 22.2 & 92.0 & 65.6 &84.6 & 70.3 & 63.7 & 59.3 & 60.2 \\
    \bottomrule%
\end{tabular}}%
\label{tab:sota-test-single}%
\end{table*}%
\begin{table*}[t]
    \centering
    \caption{\textbf{LiDAR semantic segmentation performance on nuScenes valid set}. `Ours (20\%)' utilizes only 20\% of the GT annotation, the remaining 80\% of training data being automatically annotated, to achieve a performance comparable with state-of-the-art methods. All other results are obtained from the literature, where full (100\%) GT annotation is used.
    }
    \setlength{\tabcolsep}{4.5pt}
    {\begin{tabular}{cccccccccccccccccc}%
    \toprule
    & \rotatebox[origin=l]{90}{{\textbf{mIoU}}} & \rotatebox[origin=l]{90}{{barrier}} & \rotatebox[origin=l]{90}{{bicycle}} & \rotatebox[origin=l]{90}{{bus}} & \rotatebox[origin=l]{90}{{car}}  & \rotatebox[origin=l]{90}{{construction}} & \rotatebox[origin=l]{90}{{motorcycle}} & \rotatebox[origin=l]{90}{{pedestrian}} & \rotatebox[origin=l]{90}{{trafficcone}}  & \rotatebox[origin=l]{90}{{trailer}} & \rotatebox[origin=l]{90}{{truck}}  & \rotatebox[origin=l]{90}{{driveable}} & \rotatebox[origin=l]{90}{{other}}  & \rotatebox[origin=l]{90}{{sidewalk}} & \rotatebox[origin=l]{90}{{terrain}} & \rotatebox[origin=l]{90}{{manmade}} & \rotatebox[origin=l]{90}{{vegetation}} \\
    \midrule
    (AF)2-S3Net~\cite{(AF)2-S3Net} & \color{Green}{62.2} & 60.3 & 12.6 & 82.3 & 80.0 & 20.1 & 62.0 & 59.0 & 49.0 & 42.2 & 67.4 & 94.2 & 68.0 & 64.1 & 68.6 & 82.9 & 82.4 \\
    RangeNet++~\cite{Milioto2019RangeNetF}  & \color{Green}{65.5} & 66.0 & 21.3 & 77.2 & 80.9 & 30.2 & 66.8 & 69.6 & 52.1 & 54.2 & 72.3 & 94.1 & 66.6 & 63.5 & 70.1 & 83.1 & 79.8 \\
    PolarNet~\cite{zhang2020polarnet} & \color{Green}{71.0} & 74.7 & 28.2 & 85.3 & 90.9 & 35.1 & 77.5 & 71.3 & 58.8 & 57.4 & 76.1 & 96.5 & 71.1 & 74.7 & 74.0 & 87.3 & 85.7 \\
    PVD~\cite{point-to-Voxel} & \color{Red}{76.0} & 76.2 & 40.0 & 90.2 & 94.0 & 50.9 & 77.4 & 78.8 & 64.7 & 62.0 & 84.1 & 96.6 & 71.4 & 76.4 & 76.3 & 90.3 & 86.9 \\
    CylAsy3D~\cite{zhu2021cylindrical} & \color{Red}{76.1} & 76.4 & 40.3 & 91.2 & 93.8 & 51.3 & 78.0 & 78.9 & 64.9 & 62.1 & 84.4 & 96.8 & 71.6 & 76.4 & 75.4 & 90.5 & 87.4 \\
    \midrule
    Cylinder3D (20\%) & 62.0 & 66.4 & 13.8 &  74.7 &  82.8 & 16.1 & 52.1 &  63.3 & 48.4 & 39.3 & 71.8 & 95.0 & 61.6 & 68.1 & 71.1 &  85.0 & 82.6 \\
    \rowcolor{gray!30}Ours (20\%) & 71.8 & 73.8 & 29.3 & 85.0 & 90.4 & 41.6 & 73.6 & 74.1 & 61.1 & 54.9 & 78.0 & 96.1 & 70.8 & 73.3 & 73.9 & 87.1 & 85.4 \\
    \bottomrule
\end{tabular}}%
    \label{tab:ss-nuScenes}
\end{table*}
\begin{table*}[t]
    \caption{\textbf{LiDAR semantic segmentation performance on SemanticKITTI \textit{multi-scan} test set}. Moving object classes are prefixed with `mv'; N.B., our model fails to segment `moving-truck' and `moving-other' objects as there are no examples of such categories in the 20\% labeled split. This is the limitation of the data split.
    }
    \centering%
    \setlength{\tabcolsep}{2.2pt}
    {\begin{tabular}{r@{\hskip 0.6em}r@{\hskip 0.4em}ccccccccccccccccccccccc}%
    \toprule%
         & \rotatebox[origin=l]{90}{{\textbf{mIoU}}} & {\rotatebox[origin=l]{90}{car}} & {\rotatebox[origin=l]{90}{bicycle}} & {\rotatebox[origin=l]{90}{motorcycle}} & {\rotatebox[origin=l]{90}{truck}} & {\rotatebox[origin=l]{90}{o-vehicle}} & {\rotatebox[origin=l]{90}{person}} & 
         {\rotatebox[origin=l]{90}{road}} & {\rotatebox[origin=l]{90}{parking}} & {\rotatebox[origin=l]{90}{sidewalk}} & {\rotatebox[origin=l]{90}{o-ground}} & {\rotatebox[origin=l]{90}{building}} & {\rotatebox[origin=l]{90}{fence}}& {\rotatebox[origin=l]{90}{vegetation}} & {\rotatebox[origin=l]{90}{trunk}} & {\rotatebox[origin=l]{90}{terrain}} & {\rotatebox[origin=l]{90}{pole}} & {\rotatebox[origin=l]{90}{traffic-sign}} & {\rotatebox[origin=l]{90}{mv-car}}& {\rotatebox[origin=l]{90}{mv-truck}}& {\rotatebox[origin=l]{90}{mv-other}}& {\rotatebox[origin=l]{90}{mv-person}}& {\rotatebox[origin=l]{90}{mv-biclist}}& {\rotatebox[origin=l]{90}{mv-motor}}\\
    \midrule%
      DarkNet53~\cite{SemanticKitti2019iccv}  & \textcolor{Green}{41.6}  & 84.1  & 30.4  & 32.9  & 20.0  & 20.7  & 7.5   
      & 91.6  & 64.9  & 75.3  & 27.5  & 85.2  & 56.5  & 78.4  & 50.7  & 64.8  & 38.1  & 53.3  & 61.5  & 14.1  & 15.2  & 0.2  & 28.9  & 37.8 \\
      SqueezesSegv3~\cite{SpSequenceNet} & \textcolor{Green}{43.1}  & 88.5  & 24.0  & 26.2  & 29.2  & 22.7  & 6.3  
      & 90.1  & 57.6  & 73.9  & 27.1  & 91.2  & 66.8  & 84.0  & 66.0  & 65.7  & 50.8  & 48.7  & 53.2  & 41.2  & 26.2  & 36.2  & 2.3  & 0.1 \\
      KPConv~\cite{thomas2019kpconv}  &  \textcolor{red}{51.2}  & 93.7  & 44.9  & 47.2  & 42.5  & 38.6  & 21.6 
      & 86.5  & 58.4  & 70.5  & 26.7  & 90.8  & 64.5  & 84.6  & 70.3  & 66.0  & 57.0  & 53.9  & 69.4  & 0.5  & 0.5  & 67.5  & 67.4  & 47.2 \\
      CylAsy3D~\cite{zhu2021cylindrical}  & \textcolor{red}{51.5}  & 93.8  & 67.6  & 63.3  & 41.2  & 37.6  & 12.9 
      & 90.4  & 66.3  & 74.9  & 32.1  & 92.4  & 65.8  & 85.4  & 72.8  & 68.1  & 62.6  & 61.3  & 68.1  & 0.0  & 0.1  & 63.1  & 60.0  & 0.4 \\ 
      \midrule
      Cylinder3D (20\%) & 42.1 & 89.4 & 35.2 & 22.9 & 16.3 & 15.9 & 11.6
      & 88.1 & 53.9 & 69.2 & 12.6 & 88.6 & 56.8 & 83.2 & 65.7 & 61.3 & 53.2 & 59.2 & 65.8 & 0.0 & 0.0 & 43.3 & 47.9 & 12.8 \\
      \rowcolor{gray!30}Ours (20\%) & 47.2 & 93.0 & 45.3 & 35.7 & 27.4 & 19.4 & 14.4
      & 90.5 & 61.4 & 75.0 & 15.6 & 91.3 & 62.1 & 83.3 & 69.3 & 64.0 & 59.7 & 63.6 & 77.4 & 0.0 & 0.0 & 64.0 & 57.5 & 9.5 \\
      \bottomrule%
    \end{tabular}}%
    \label{tab:sota-multi-scan-test}%
\end{table*}%

\subsection{Datasets}
\label{sec:dataset}
We evaluate our approach on the Argoverse dataset~\cite{Argoverse} for 3D \textit{vehicle} object detection, and SemanticKITTI~\cite{SemanticKitti2019iccv} and nuScenes~\cite{nuScenes_2020} for 3D semantic segmentation. 
\textbf{Argoverse} provides a large collection of LiDAR sequences with 3D bounding-box labels, from which we utilize the first $10\%$ of human-labeled sequences ($\Tl$) and $90\%$ being gathered without annotation in $\Tu$ for pseudo-labeling. \textbf{SemanticKITTI} provides a large-scale set of driving-scene sequences for 3D semantic segmentation. It consists of 22 sequences that split from 00 to 10 for training (08 reserved for validation) and 11 to 21 for testing. The dataset has two challenges, i.e., \emph{single-scan} with 19 class categories and \emph{multi-scan} with 25 class categories, including 19 from single-scan and six moving-object categories. \textbf{nuScenes} contains 1000 scenes with a great diversity of urban traffic and weather conditions. It officially divides the data into 700/150/150 scenes for train/val/test.
For our experiment, we cut each sequence into two parts, the first 20\% for the human-labeled set $\Tl$ and the latter 80\% for the unlabeled set $\Tu$.

\subsection{3D Multi-Class Semantic Segmentation}
We train a student model on pseudo-labels generated by the concordance of teachers. Here, we utilize Cylinder3D and the output class probabilities for all individual scan points are treated with our confidence-guided criterion (Eqs.\,\ref{eq:pseudo-label},\,\ref{eq:computed-weight}).
Training is done by optimizing the cross-entropy loss and the Lovasz-softmax loss~\cite{berman2018lovasz} weighted by our confidence-guided criterion, as in Eq.\,\ref{eq:mean-loss}. The standard mean Intersection over Union (mIoU) metric is used for evaluation.

\begin{figure*}[t]
    \centering
    \begin{subfigure}[]{0.49\textwidth}
      \includegraphics[width=\textwidth]{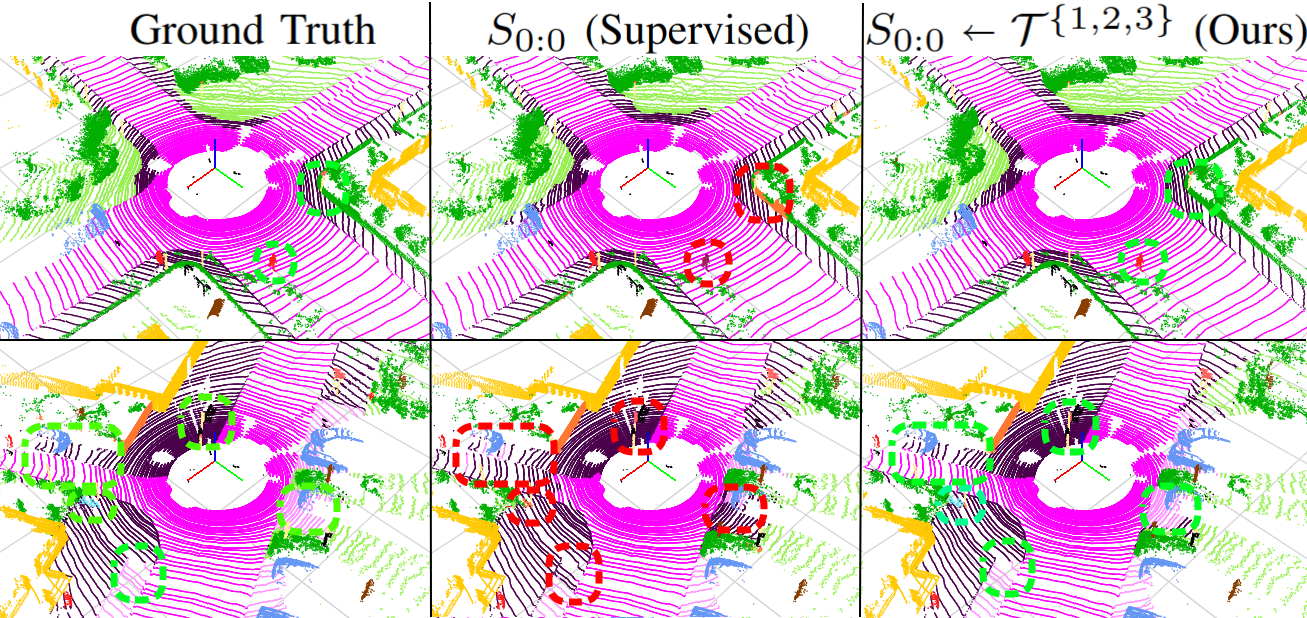}
        \caption{
        A scene from SemanticKITTI with its GT semantic segmentation (1st column) showing that our method $\stud_{\shortminus 0:0} \!\shortleftarrow\! \concordance^{\{1,2,3\}}$ (3rd column) can segment person~\crule[person-color]{0.14cm}{0.14cm}, side-walks~\crule[side-walks-color]{0.14cm}{0.14cm}, bicycle~\crule[bicycle-color]{0.14cm}{0.14cm} and drivable areas~\crule[road-color]{0.14cm}{0.14cm} better than the supervised baseline based on Cylinder3D (2nd column).
        Correct and incorrect segmentations are indicated by green and red circles, respectively. Both methods utilize 20\% of human-labeled data.
        }
        \label{fig:ss-result}
    \end{subfigure}
    ~
    \begin{subfigure}[]{0.49\textwidth}
        \includegraphics[width=\textwidth]{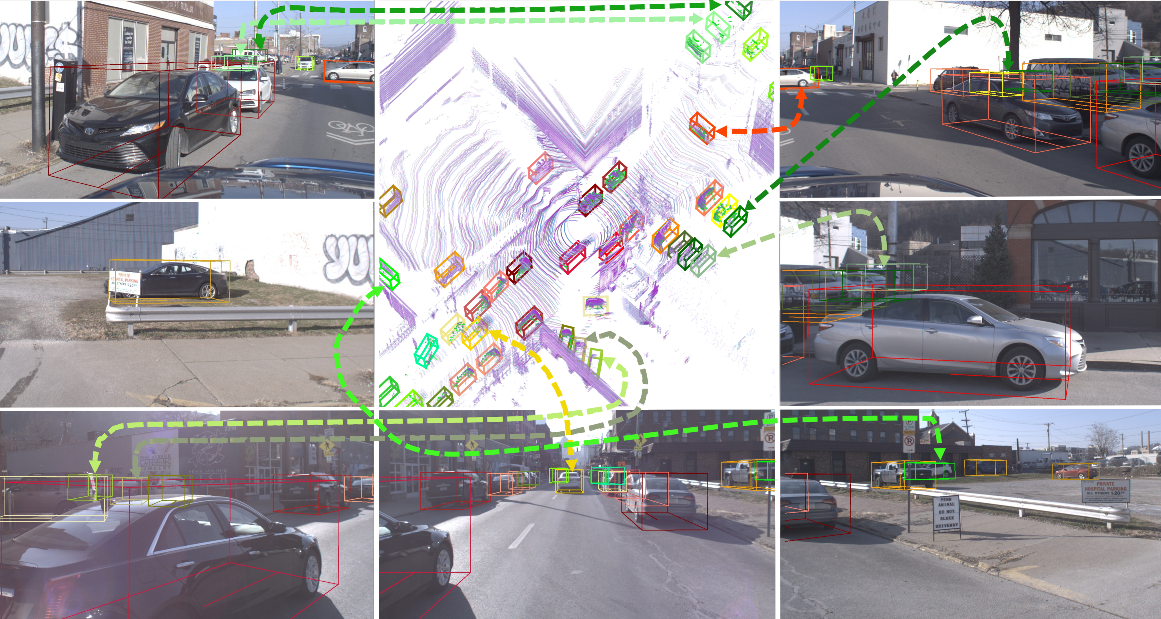}
    \caption{
    A scene from Argoverse showing that student $\stud_{\shortminus 5:0} \!\shortleftarrow\! \mathcal{T}^{\{3,4,5\}}$ can be robust to severe occlusions. Some of the correctly detected boxes do not contain GT vehicle points since they correspond to vehicles occluded in the current frame and partially occluded in the past frames, as can be observed in the ring of camera images.
    }
    \label{fig:lidar-rgb}
    \end{subfigure}
    
    \caption{\textbf{Qualitative results}. Examples of (a) 3D semantic segmentation and (b) 3D vehicle detection.}
    \label{fig:my_label}
\end{figure*}

\smallskip\noindent\textbf{Single-scan semantic segmentation.}
In this experiment, we compare the results of our method with fully-supervised state-of-the-art LiDAR segmentation on SemanticKITTI single-scan test set and nuScenes validation set. Further, we present some qualitative results in Fig.\,\ref{fig:ss-result}, which show that our model helps to improve the segmentation quality as compared to its supervised-only counterparts. 
As shown in Table \ref{tab:sota-test-single}, our method $\stud_{0:0} \!\shortleftarrow\! \concordance^{\{1,2,3\}}$ trained with only $20\%$ of ground truth (GT) and $80\%$ of pseudo-labeled outperforms all methods based on 3D-to-2D projection with fully-annotated training data \cite{xu2020squeezesegv3,Milioto2019RangeNetF,zhang2020polarnet}.
Moreover, our method shows comparable results to voxel partition and 3D convolution-based methods, including fully-supervised Cylinder3D. 
We made a similar comparison with the fully-supervised state-of-the-art methods on nuScenes validation split.
Our method $\stud_{0:0}\shortleftarrow\concordance^{\{1,2,3\}}$ trained with only $20\%$ of GT and $80\%$ of pseudo-labels outperforms some of the fully-supervised models, see Table~\ref{tab:ss-nuScenes}.

\smallskip\noindent\textbf{Multi-scan semantic segmentation.} Compared to the single-scan set-up, the multi-scan segmentation in SemanticKITTI has six more categories accounting for moving objects (\textit{car}, \textit{truck}, \textit{other-vehicle}, \textit{person}, \textit{bicyclist} and \textit{motorcyclist}). In this experiment, all methods utilize multiple input point clouds. In Table\,\ref{tab:sota-multi-scan-test}, we show that our method, with only $20\%$ of human-labeled training data, outperforms methods that use full manual annotations, namely, DarkNet53~\cite{SemanticKitti2019iccv} and SqueezesSegv3~\cite{SpSequenceNet}, and is on par with KPConv~\cite{thomas2019kpconv} and CylAsy3D~\cite{zhu2021cylindrical}. Our method outperforms all others on the \textit{moving-car} and \textit{traffic-sign} categories.
\begin{table}[t]
    \centering
    \caption{\textbf{Semi-supervised learning on SemanticKITTI validation set}. Performance in mIoU (\%).
    ‘Guided-Point-SSL’ denotes \cite{jiang2021guided} semi-supervised models;
     ‘$\stud_{0:0} \!\shortleftarrow\! \concordance^{\{1,2,3\}}$’ denotes our approach with distillation from the concordance of teachers.}
    \begin{tabular}{c@{\hskip 1em}ccccc}
        \toprule
        &  	& Labeled data \\				
        \cmidrule{2-4}
        method  & 20\%	& 30\%	& 40\%  \\
        \midrule
        Guided Point SSL \cite{jiang2021guided} &	58.8 &	59.4	& 59.9  \\
        \rowcolor{gray!30}$\stud_{0:0} \!\shortleftarrow\! \concordance^{\{1,2,3\}}$ (Ours) & \textbf{59.9}	& \textbf{60.7}	& \textbf{62.2} \\
        \bottomrule
    \end{tabular}
    \label{tab:sota-ssl}
\end{table}
%
%
%

\smallskip\noindent\textbf{Comparison to state-of-the-art semi-supervised method.} To further assess the merit of our approach, we compare it to the most recent semi-supervised segmentation work, Guided-Point-SSL~\cite{jiang2021guided} on the SemanticKITTI validation set. As shown in Table\,\ref{tab:sota-ssl}, our method learned from Concordance of teachers, $\stud_{0:0}\!\shortleftarrow\!\concordance^{\{1,2,3\}}$, outperforms Guided-Point-SSL with $20\%$, $30\%$ and $40\%$ labeled data by 1.1, 1.3 and 2.3 mIoU points respectively.
%
%
%
%
\subsection{3D Object Detection} 
The models are trained in the same way as described in PointRCNN \cite{shi2019pointrcnn}, except for the confidence-guided criterion and the usage of multiple frames at the input.
\begin{table}[t]
    \centering
    \caption{\textbf{Results of 3D object detection}. Detection performance (AP percentage) of proposed students trained with 10\% human-labeled training data and of oracle model trained with full (100\%) labeled data on Argoverse validation set. `*' indicates reimplementation into our multi-frame PointRCCN architecture.}
    \setlength{\tabcolsep}{4.5pt}
    {\begin{tabular}{cccc}%
    \toprule
     $\stud_{\shortminus 5:0}$~\cite{shi2019pointrcnn}* &      $\stud_{\shortminus 5:0} \shortleftarrow \teach_{\shortminus 5:0}$~\cite{SESS_2020_CVPR}* &$\stud_{\shortminus 5:0}\!\shortleftarrow \!\concordance^{\{3,4,5\}}$ (Ours) & Oracle \\ 
    \midrule
    51.1 & 56.9 & {58.3} & 62.6 \\
    \bottomrule
    \end{tabular}}
    \label{tab:oracle}
\end{table}
In Table\,\ref{tab:oracle}, we compare the proposed method to the baseline $\stud_{\shortminus 5:0}$~\cite{shi2019pointrcnn} and to the Mean-Teacher framework (MT) $\stud_{\shortminus5:0} \!\shortleftarrow\! \teach_{\shortminus5:0}$~\cite{SESS_2020_CVPR}. To reach a fair comparison, we have re-implemented the MT~\cite{SESS_2020_CVPR} into our architecture and used the classification and regression branches of the original PointRCNN loss function instead of the MT~\cite{SESS_2020_CVPR} consistency loss. The proposed method $\stud_{\shortminus 5:0} \!\shortleftarrow\! \concordance^{\{3,4,5\}}$ learned from the concordance of teachers achieves 58.3\,AP when trained with 10\% human-labeled training data, outperforming the baseline~\cite{shi2019pointrcnn} by 7.2 AP and the MT~\cite{SESS_2020_CVPR} by 1.4 AP. Moreover, it closes 62.6\% of the gap between the baseline~\cite{shi2019pointrcnn} and the `Oracle' (the model $\stud_{\shortminus 5:0}$ trained with 100\% GT labeled training data).

\subsection{Implementation Details}\label{sec:implementation_details}
We trained models with an ADAM optimizer with a learning rate of 0.001 for 40 epochs on semantic segmentation and 200 and 50 epochs of the RPN and RCNN branches of the object detection model, respectively, using 4 Nvidia A100 GPUs running for ~3 days of training for each task. 
In the object detection task, we subsample each point cloud to 16,384 points from each frame as inputs to the model. We have used three set-abstraction layers $\phi$ with sizes 4096, 1024, and 128 for our multi-scale time-aware grouping to subsample points into groups. We have used $\lambda = 0.1$ in our experiments.

\section{Ablation Studies} 
\smallskip\noindent\textbf{Ablation on temporal diversity of teachers}. 
\begin{table}[t]%
    \centering%
    \caption{\textbf{Effect of temporal diversity of teachers.} The student trained using concordance of teachers from different temporal ranges outperforms the one trained from the same temporal range but with different initialization in both (a) 3D object detection (Argoverse validation set) and (b) semantic segmentation (SemanticKITTI validation set).}%
    \begin{subtable}{.4\linewidth}
      \centering
      \caption{Object detection (AP)}
      \begin{tabular}{ccc}
      \toprule
        $\stud_{\shortminus 3:0} \!\shortleftarrow\! \mathcal{E}^{\{3,3\}}$  & 54.3    \\
        $\stud_{\shortminus 3:0} \!\shortleftarrow\! \concordance^{\{2,3\}}$ & \bf{55.6} \\
        \midrule
        $\stud_{\shortminus 4:0} \!\shortleftarrow\! \mathcal{E}^{\{4,4\}}$  & 57.2    \\
        $\stud_{\shortminus 4:0} \!\shortleftarrow\! \concordance^{\{3,4\}}$ & \bf{57.7} \\
        \midrule
        $\stud_{\shortminus 5:0} \!\shortleftarrow\! \mathcal{E}^{\{5,5\}}$  & 58.0    \\
        $\stud_{\shortminus 5:0} \!\shortleftarrow\! \concordance^{\{4,5\}}$ & \bf{58.2} \\
        \bottomrule
       \end{tabular}
       \label{tab:od-temporal}
    \end{subtable}
    ~
    \begin{subtable}{.5\linewidth}
      \centering
       \caption{Semantic seg. (mIoU)}
    	\begin{tabular}{ccc}
        \toprule
        $\stud_{\shortminus 1:0} \!\shortleftarrow\! \mathcal{E}^{\{1,1,1\}}$ & 59.5 \\
        $\stud_{\shortminus 1:0} \!\shortleftarrow\! \concordance^{\{1,2,3\}}$ & \bf{60.6} \\
        \midrule
        $\stud_{\shortminus 2:0} \!\shortleftarrow\! \mathcal{E}^{\{2,2,2\}}$ & 59.9 \\
        $\stud_{\shortminus 2:0} \!\shortleftarrow\! \concordance^{\{1,2,3\}}$ & \bf{60.6} \\
        \midrule
        $\stud_{\shortminus 3:0} \!\shortleftarrow\! \mathcal{E}^{\{3,3,3\}}$ & 60.0 \\
        $\stud_{\shortminus 3:0} \!\shortleftarrow\! \concordance^{\{1,2,3\}}$ & \bf{60.9} \\
        \bottomrule
    \end{tabular}
    \label{tab:temporal-ss}
    \end{subtable}
    \label{tab:temporal-diversity}
\end{table}
We show that the temporal diversity among teachers overperforms teachers with the same temporal range but with various training initializations. Following the Concordance notation, a set of teachers with the same temporal range is denoted $\mathcal{E}^{\{n, \dots, n\}} = {\{T^1_{\shortminus n,n}, \dots, T^K_{\shortminus n,n}\}}$, where each randomly initialized teacher $T^k_{\shortminus n,n}$ is operating in \emph{the same} temporal range as others. As shown in Table\,\ref{tab:temporal-diversity}, the student trained by  
teachers from different temporal ranges outperforms one
trained by teachers on the same temporal range, in both 3D object detection and 3D semantic segmentation.

\begin{figure}[t]%
    \centering%
    \includegraphics[width=0.9\linewidth, height=2.9cm]{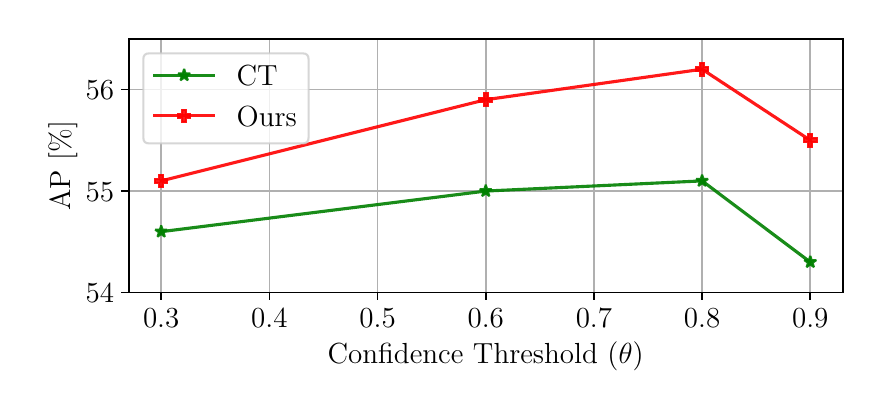}
    \captionof{figure}{\textbf{Ablation on PL's selection strategy in Argoverse object detection}.  Performance as a function of threshold $\theta$. `Ours' is the proposed  confidence-guided criterion and `CT' the standard confidence-based threshold~\cite{DBLP:pseudo_objdet,Zhou_2021_CVPR,wang20203dioumatch}.}
		\label{fig:ablation}%
\end{figure}

\smallskip\noindent\textbf{Selection of pseudo-labels.~}This ablation study demonstrates the benefits of the proposed confidence-guided criterion. The standard baselines here are tuning a single confidence-based threshold (CT) for all pseudo-labels~\cite{DBLP:pseudo_objdet,Zhou_2021_CVPR,wang20203dioumatch}. Models with our proposed selection criterion outperform the CT across multiple confidence thresholds, see Fig.\,\ref{fig:ablation}.
\begin{figure}[t]
    \centering   \includegraphics[width=.93\linewidth,height=0.14\textheight]{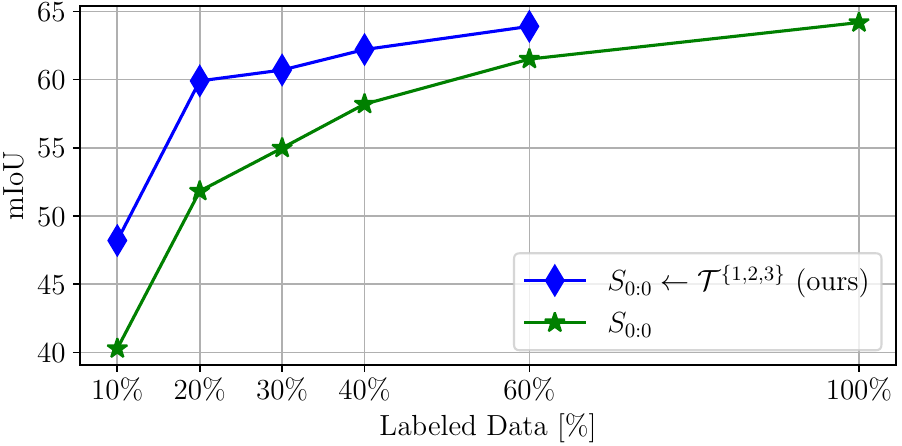}
    \caption{\textbf{Impact of labeling proporiton}. Semantic segmentation performance on SemanticKITTI validation set as a function of labeled data proportion size.}
    \label{fig:labeled-pseudo-labeled}
\end{figure}

\smallskip\noindent\textbf{Effect of labeled and pseudo-labeled dataset ratio.~}We performed an ablation study to understand the effect of labeled vs. pseudo-labeled training data. We have used the same architecture setup, only varying the amount of labeled and pseudo-labeled data, $|\Tl| =$ 10, 20, 30, 40, 60, and 100\% of training data.
As shown in Fig.\,\ref{fig:labeled-pseudo-labeled}, the gain increases significantly by $\sim{}10$ mIoU when the model uses $|\Tl|=20\%$ of training data compared to $|\Tl|=10\%$. However, the relative gain decreases as the number of labeled data increases to 30 and 40\%. This trend shows that the size ratio between $\Tl$ and $\Tp$ should be carefully set to achieve adequate performance with the smallest amount of labeled data possible. Moreover, the proposed method, when it uses $|\Tl|=60\%$ of training data (and $|\Tp|=40\%$), reaches the performance of the fully-supervised baseline model $S_{0:0}$ which is Cylinder3D~\cite{zhou2020cylinder3d} trained with $100\%$ of labeled training data; it can be observed by comparing the top-right ending points on the blue and green lines.
\begin{table}[t]
    \centering
    \caption{\textbf{Comparision to other teacher-student methods}. All methods use $\stud_{0:0}$, are trained with 20\% of labeled data and are evaluated on the SemanticKITTI validation set.}
    \begin{tabular}{cccc}
        \toprule
        methods & mIoU [\%] \\
        \midrule
        Cylinder3D + KD \cite{Know_distill} & 54.8  \\
        Cylinder3D + EN \cite{rokach2019ensemble} & 56.0 \\
        \rowcolor{gray!30}Cylinder3D + Ours & 59.9\\
        \bottomrule
    \end{tabular}
    \label{tab:other-teacher-student}
\end{table}

\smallskip\noindent\textbf{Comparision to other teacher-student frameworks.} We compare in Table\,\ref{tab:other-teacher-student} our method with other teacher-student approaches such as knowledge distillation (KD)~\cite{Know_distill}, 
and Ensemble (EN)~\cite{rokach2019ensemble}. We report a comparison using the Cylinder3D $S_{0:0}$ model trained with the methods above using the hyperparameters from Section~\ref{sec:implementation_details}. All methods are trained with 20\% labeled data. As shown in Table\,\ref{tab:other-teacher-student}, the proposed method significantly outperforms  KD~\cite{Know_distill} and EN~\cite{rokach2019ensemble} baseline teacher-student methods.  

\begin{figure}[t]
    \centering
    \includegraphics[width=0.99\linewidth]{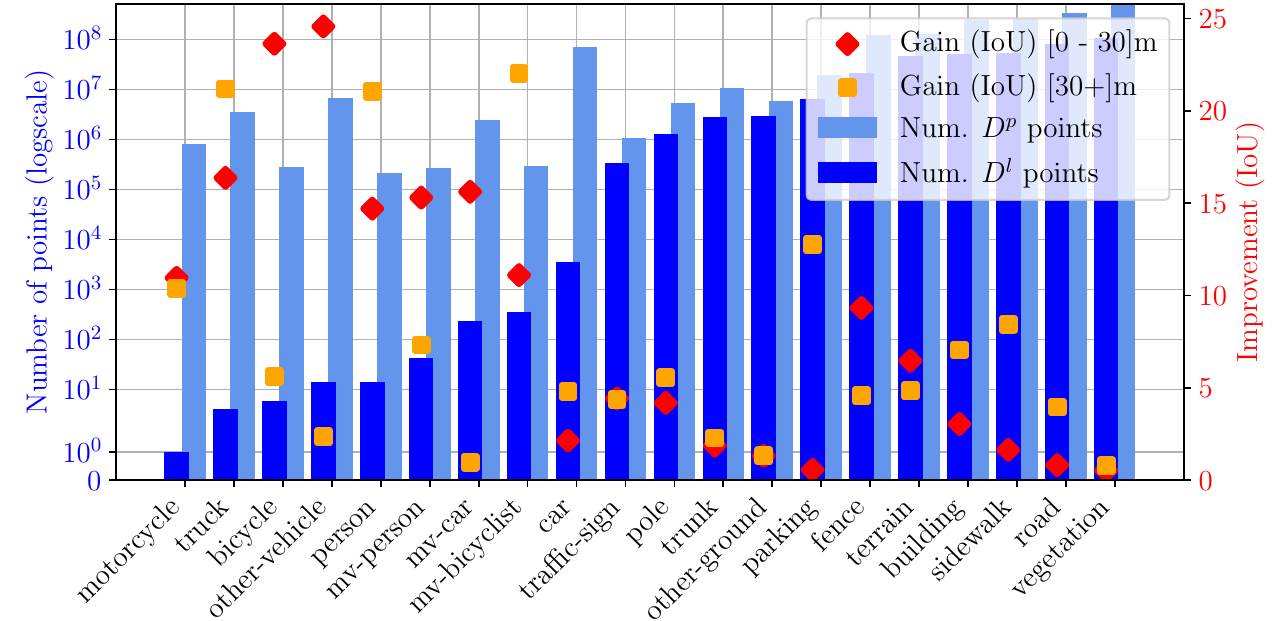}
    \caption{\textbf{Distance-based improvement in class IoUs on SemanticKITTI.} Classes are sorted according to the number of associated labeled training points in $\Tl$ and visualized together with pseudo-labeled points ($\Tp$).
    We show relative IoU gain within a 30-meter distance from the ego-vehicle ([0-30]m) and at a further distance ([30+]m).
    Moving objects are prefixed with ‘mv’; N.B.: we have omitted classes that do not have points in the labeled split $\Tl$.}
    \label{fig:calss-iprovement}
\end{figure}

\smallskip\noindent\textbf{Distant and close objects.~} We investigate how multi-scan segmentation is affected by the distance of the points to the ego-vehicle and the number of labeled and pseudo-labeled points for each object class. We compare our method to the baseline \cite{zhou2020cylinder3d} on the SemanticKITTI validation set to show the relative gain. Both models are trained with 20\% labeled training data. As shown in Fig.\,\ref{fig:calss-iprovement}, a notable performance boost is observed for rarely-appearing classes, especially within 30 meters distance from the ego-vehicle.

\smallskip\noindent\textbf{Limitations.~} The SemanticKITTI dataset has a huge data imbalance, and we perform the $\Tl$ and $\Tu$ split without any relevance to the number of points per object class.
We observe cases where no points belong to a specific object category in the labeled set $\Tl$ resulting in teachers' inability to recognize a particular class, see `mv-truck' and `mv-other' in Table\,\ref{tab:sota-multi-scan-test}. One should ensure that all the object categories are present in the labeled set $\Tl$.
Secondly, in object detection, since we estimate only objects in the reference frame, we sometimes observe false \textit{false positives}, i.e., detections that are correct but missing in the GT annotation due to no points in the reference frame. The model learns point features from different times and estimate vehicle position in the reference frame. We did not address this issue in evaluating the object detection task.

\section{Conclusion}
\label{sec:conclusion}
We propose a novel pseudo-labeling framework that leverages spatio-temporal information from unlabeled sequences of point clouds. We demonstrate its merit in two 3D perception tasks on publicly available datasets. 
The reported performance gains stem from (i) A better selection of the final pseudo-labels via the concordance of multiple teachers operating at different temporal ranges; (ii) A novel pseudo-label confidence-guided criterion.
Thanks to the privileged information available in the different temporal ranges, the Concordance of teachers delivers strong pseudo-labeled samples. 
Using manual labeling of only 20\% of training data, our method achieves state-of-the-art performance in semi-supervised 3D semantic segmentation and competes even with methods that use the full set of labels on this task.
By the nature of our pseudo-labeling framework, the proposed approach is complementary to other techniques that use sequential data, and can thus be combined with them to further boost the performance. performance.

\section{Acknowledgment}
This work was supported in part by OP VVV MEYS funded project CZ.02.1.01/0.0/0.0/16\_019/0000765 ``Research Center for Informatics'', and by 
CTU Prague Project SGS22/111/OHK3/2T/13. K. Zimmermann acknowledges CSF Project 20-29531S. The authors want to thank Valeo for its support.\\

\bibliographystyle{plain}

\begin{thebibliography}{10}


\bibitem{SemanticKitti2019iccv}
J.~Behley, M.~Garbade, A.~Milioto, J.~Quenzel, S.~Behnke, C.~Stachniss, and
  J.~Gall.
\newblock {SemanticKITTI: A Dataset for Semantic Scene Understanding of LiDAR
  Sequences}.
\newblock In {\em Proc. of the IEEE/CVF International Conf.~on Computer Vision
  (ICCV)}, 2019.

\bibitem{berman2018lovasz}
Maxim Berman, Amal~Rannen Triki, and Matthew~B Blaschko.
\newblock The lov{\'a}sz-softmax loss: A tractable surrogate for the
  optimization of the intersection-over-union measure in neural networks.
\newblock In {\em CVPR}, 2018.

\bibitem{waymo_pseudo_labelling-short}
Benjamin Caine et~al.
\newblock Pseudo-labeling for scalable 3d object detection.
\newblock {\em arXiv}, abs/2103.02093, 2021.

\bibitem{Argoverse}
Ming-Fang Chang, John~W Lambert, Patsorn Sangkloy, Jagjeet Singh, Slawomir Bak,
  Andrew Hartnett, De~Wang, Peter Carr, Simon Lucey, Deva Ramanan, and James
  Hays.
\newblock Argoverse: 3{D} tracking and forecasting with rich maps.
\newblock In {\em IEEE/CVF Conference on Computer Vision and Pattern
  Recognition (CVPR)}, 2019.

\bibitem{Knowledge_efficiancy_2019_ICCV}
Jang~Hyun Cho and Bharath Hariharan.
\newblock On the efficacy of knowledge distillation.
\newblock In {\em IEEE/CVF International Conference on Computer Vision (ICCV)},
  October 2019.

\bibitem{Minknet}
Christopher~B. Choy, JunYoung Gwak, and Silvio Savarese.
\newblock 4d spatio-temporal convnets: Minkowski convolutional neural networks.
\newblock In {\em {IEEE} Conference on Computer Vision and Pattern Recognition,
  {CVPR} 2019, Long Beach, CA, USA, June 16-20, 2019}, pages 3075--3084.
  Computer Vision Foundation / {IEEE}, 2019.

\bibitem{cortinhal2021salsanext}
Tiago Cortinhal, George Tzelepi, and Eren Aksoy.
\newblock Salsanext: Fast, uncertainty-aware semantic segmentation of lidar
  point clouds for autonomous driving.
\newblock In {\em 15th International Symposium ISVC 2020}.

\bibitem{gerdzhev2021tornado}
Martin Gerdzhev, Ryan Razani, Ehsan Taghavi, and Liu Bingbing.
\newblock Tornado-net: multiview total variation semantic segmentation with
  diamond inception module.
\newblock In {\em 2021 IEEE International Conference on Robotics and Automation
  (ICRA)}, pages 9543--9549. IEEE, 2021.

\bibitem{Knowledge_orig}
Geoffrey Hinton, Oriol Vinyals, Jeff Dean, et~al.
\newblock Distilling the knowledge in a neural network.
\newblock In {\em NIPS Deep Learning and Representation Learning Workshop},
  2015.

\bibitem{what_you_see}
Peiyun Hu, Jason Ziglar, David Held, and Deva Ramanan.
\newblock What you see is what you get: Exploiting visibility for 3d object
  detection.
\newblock In {\em CVPR 2020}, pages 10998--11006, 06 2020.

\bibitem{hu2020randla}
Qingyong Hu, Bo~Yang, Linhai Xie, Stefano Rosa, Yulan Guo, Zhihua Wang, Niki
  Trigoni, and Andrew Markham.
\newblock Randla-net: Efficient semantic segmentation of large-scale point
  clouds.
\newblock In {\em Proc. of the IEEE/CVF Conference on Computer Vision and
  Pattern Recognition}, pages 11108--11117, 2020.

\bibitem{jiang2021guided}
Li~Jiang, Shaoshuai Shi, Zhuotao Tian, Xin Lai, Shu Liu, Chi-Wing Fu, and Jiaya
  Jia.
\newblock Guided point contrastive learning for semi-supervised point cloud
  semantic segmentation.
\newblock In {\em Proc. of the IEEE/CVF International Conference on Computer
  Vision}, pages 6423--6432, 2021.

\bibitem{Dlupi}
John Lambert, Ozan Sener, and Silvio Savarese.
\newblock Deep learning under privileged information using heteroscedastic
  dropout.
\newblock In {\em IEEE/CVF Conference on Computer Vision and Pattern
  Recognition (CVPR)}, pages 8886--8895, 2018.

\bibitem{Lee2013PseudoLabelT}
D.~Lee.
\newblock Pseudo-label : The simple and efficient semi-supervised learning
  method for deep neural networks.
\newblock In {\em International Conference on Machine Learning (ICML). Workshop
  Challenges in Representation Learning}, 2013.

\bibitem{li2007optimol}
Li{-}Jia Li, Gang Wang, and Fei{-}Fei Li.
\newblock {OPTIMOL:} automatic online picture collection via incremental model
  learning.
\newblock In {\em {IEEE/CVF} Conference on Computer Vision and Pattern
  Recognition {(CVPR})}. {IEEE} Computer Society, 2007.

\bibitem{Li2020SelfLoopUA}
Yuexiang Li, J.~Chen, Xinpeng Xie, Kai Ma, and Y.~Zheng.
\newblock Self-loop uncertainty: A novel pseudo-label for semi-supervised
  medical image segmentation.
\newblock {\em The Medical Image Computing and Computer Assisted Intervetion,
  (MICCAI)}, pages 614--623, 09 2020.

\bibitem{Knowledge_flow}
I~Liu et~al.
\newblock Knowledge flow: Improve upon your teachers.
\newblock In {\em 7th International Conference on Learning Representations,
  (ICLR)}, 2019.

\bibitem{Meteornet}
Xingyu Liu et~al.
\newblock Meteornet: Deep learning on dynamic 3d point cloud sequences.
\newblock In {\em ICCV}, 2019.

\bibitem{DBLP:pseudo_objdet}
Yen{-}Cheng Liu, Chih{-}Yao Ma, Zijian He, Chia{-}Wen Kuo, Kan Chen, Peizhao
  Zhang, Bichen Wu, Zsolt Kira, and Peter Vajda.
\newblock Unbiased teacher for semi-supervised object detection.
\newblock In {\em International Conference on Learning Representations,
  (ICLR)}, 2021.

\bibitem{Milioto2019RangeNetF}
Andres Milioto, Ignacio Vizzo, Jens Behley, and C.~Stachniss.
\newblock Rangenet ++: Fast and accurate lidar semantic segmentation.
\newblock {\em 2019 IEEE/RSJ International Conference on Intelligent Robots and
  Systems (IROS)}, pages 4213--4220, 2019.

\bibitem{Teacher_assistant}
Seyed-Iman Mirzadeh, Mehrdad Farajtabar, Ang Li, and Hassan Ghasemzadeh.
\newblock Improved knowledge distillation via teacher assistant: Bridging the
  gap between student and teacher.
\newblock In {\em 34th AAAI Conference on Artificial Intelligence}, 2019.

\bibitem{off-board_waymo}
Charles~R Qi, Yin Zhou, Mahyar Najibi, Pei Sun, Khoa Vo, Boyang Deng, and
  Dragomir Anguelov.
\newblock Offboard 3d object detection from point cloud sequences.
\newblock In {\em CVPR}, 2021.

\bibitem{journals/corr/pointnet}
Charles~Ruizhongtai Qi, Hao Su, Kaichun Mo, and Leonidas~J. Guibas.
\newblock Point{N}et: Deep learning on point sets for 3{D} classification and
  segmentation.
\newblock In {\em IEEE/CVF Conference on Computer Vision and Pattern
  Recognition (CVPR)}, 2017.

\bibitem{journals/symmetry/pointnet++}
Charles~Ruizhongtai Qi, Li~Yi, Hao Su, and Leonidas~J Guibas.
\newblock Pointnet++ and three layers of features fusion for occlusion
  three-dimensional ear recognition based on one sample per person.
\newblock {\em Symmetry}, 12(1):78, 2020.

\bibitem{SpSequenceNet}
Hanyu Shi, Guosheng Lin, Hao Wang, Tzu-Yi Hung, and Zhenhua Wang.
\newblock Spsequencenet: Semantic segmentation network on 4d point clouds.
\newblock In {\em Proc. of the IEEE/CVF Conference on Computer Vision and
  Pattern Recognition (CVPR)}, June 2020.

\bibitem{shi2019pointrcnn}
Shaoshuai Shi, Xiaogang Wang, and Hongsheng Li.
\newblock Point{RCNN}: 3{D} object proposal generation and detection from point
  cloud.
\newblock In {\em IEEE/CVF Conference on Computer Vision and Pattern
  Recognition (CVPR)}, pages 770--779, 2019.

\bibitem{Shi_2018_ECCV}
Weiwei Shi, Yihong Gong, Chris Ding, Zhiheng~MaXiaoyu Tao, and Nanning Zheng.
\newblock Transductive semi-supervised deep learning using min-max features.
\newblock In {\em Proc. of the European Conference on Computer Vision (ECCV)},
  September 2018.

\bibitem{Berthelot_Consistency}
Kihyuk Sohn, David Berthelot, Nicholas Carlini, Zizhao Zhang, Han Zhang,
  Colin~A Raffel, Ekin~Dogus Cubuk, Alexey Kurakin, and Chun-Liang Li.
\newblock Fixmatch: Simplifying semi-supervised learning with consistency and
  confidence.
\newblock In {\em Advances in Neural Information Processing Systems},
  volume~33, pages 596--608. Curran Associates, Inc., 2020.

\bibitem{SegContrast2022}
L. Nunes, R. Marcuzzi, X. Chen, J. Behley, and C. Stachniss. 
\newblock Seg-Contrast: 3D point cloud feature representation learning through self-supervised segment discrimination. \newblock IEEE RAL, 2022

\bibitem{jaritz2020xmuda}
M. Jaritz, T. Vu, R. d. Charette, E. Wirbel, and P. P ́erez. 
\newblock xMUDA: Cross-modal unsupervised domain adaptation for 3d semantic segmentation. 
\newblock In {\em IEEE/CVF Conference on Computer Vision and Pattern
  Recognition (CVPR)}, 2020

\bibitem{LiDARNet}
P. Jiang and S. Saripalli. Lidarnet: 
\newblock A boundary-aware domain adaptation model for point cloud semantic segmentation. 
\newblock In {\em International conference robotics automation}, 2021.

\bibitem{tatarchenko2018tangent}
Maxim Tatarchenko, Jaesik Park, Vladlen Koltun, and Qian-Yi Zhou.
\newblock Tangent convolutions for dense prediction in 3d.
\newblock In {\em Proc. of the IEEE Conference on Computer Vision and Pattern
  Recognition}, pages 3887--3896, 2018.

\bibitem{automatic-labeling2022}
X. Chen, B. Mersch, L. Nunes, R. Marcuzzi, I. Vizzo, J. Behley, and
C. Stachniss. 
\newblock Automatic labeling to generate training data for online lidar-based moving object segmentation. 
\newblock IEEE RA-L, page 6107–6114,
2022

\bibitem{tracking-based2012}
A. Teichman and S. Thrun. 
\newblock Tracking-based semi-supervised learning.
\newblock IJRR, 31(7):804–818, 2012

\bibitem{xie2020pointcontrast}
Xie, S. and Gu, J. and Guo, D. and Qi, C. R. and Guibas, L. and Litany, O.
\newblock {Pointcontrast: Unsupervised pre-training for 3{D} point cloud understanding}.
\newblock {Europian conference in computer vision (ECCV)}, 2020.

\bibitem{Know_distill}
F. Tung and G. Mori. 
\newblock Similarity-preserving knowledge distillation. 
\newblock In ICCV, 2019

\bibitem{rokach2019ensemble}
L. Rokach. Ensemble Learning:
\newblock Pattern Classification Using Ensemble Methods. 
\newblock Series in machine perception and artificial intelligence. 2019.

\bibitem{thomas2019kpconv}
Hugues Thomas, Charles~R Qi, Jean-Emmanuel Deschaud, Beatriz Marcotegui,
  Fran{\c{c}}ois Goulette, and Leonidas~J Guibas.
\newblock Kpconv: Flexible and deformable convolution for point clouds.
\newblock In {\em Proc. of the IEEE/CVF international conference on computer
  vision}, pages 6411--6420, 2019.

\bibitem{VAPNIK2009544}
Vladimir Vapnik et~al.
\newblock A new learning paradigm: Learning using privileged information.
\newblock {\em Neural Networks}, 22(5):544--557, 2009.
\newblock Advances in Neural Networks Research: IJCNN2009.

\bibitem{Vapnik_2015}
Vladimir Vapnik and Rauf Izmailov.
\newblock Learning using privileged information: Similarity control and
  knowledge transfer.
\newblock {\em Journal of Machine Learning Research}, 16(61):2023--2049, 2015.

\bibitem{wang20203dioumatch}
He~Wang, Yezhen Cong, Or~Litany, Yue Gao, and Leonidas~J Guibas.
\newblock 3{DIoUMatch}: Leveraging {IoU} prediction for semi-supervised 3d
  object detection.
\newblock {\em arXiv preprint arXiv:2012.04355}, 2020.

\bibitem{xu2020squeezesegv3}
Chenfeng Xu, Bichen Wu, Zining Wang, Wei Zhan, Peter Vajda, Kurt Keutzer, and
  Masayoshi Tomizuka.
\newblock Squeezesegv3: Spatially-adaptive convolution for efficient
  point-cloud segmentation.
\newblock In {\em European Conference on Computer Vision}, pages 1--19.
  Springer, 2020.

\bibitem{Zhang2020DeepFF}
Feihu Zhang, Jin Fang, Benjamin~W. Wah, and Philip H.~S. Torr.
\newblock Deep fusionnet for point cloud semantic segmentation.
\newblock In {\em ECCV}, 2020.

\bibitem{zhang2020polarnet}
Yang Zhang, Zixiang Zhou, Philip David, Xiangyu Yue, Zerong Xi, Boqing Gong,
  and Hassan Foroosh.
\newblock Polarnet: An improved grid representation for online lidar point
  clouds semantic segmentation.
\newblock In {\em Proc. of the IEEE/CVF Conference on Computer Vision and
  Pattern Recognition}, pages 9601--9610, 2020.

\bibitem{SESS_2020_CVPR}
Na~Zhao, Tat-Seng Chua, and Gim~Hee Lee.
\newblock Sess: Self-ensembling semi-supervised 3d object detection.
\newblock In {\em Proc. of the IEEE/CVF Conference on Computer Vision and
  Pattern Recognition (CVPR)}, June 2020.

\bibitem{zhou2020cylinder3d}
Hui Zhou, Xinge Zhu, Xiao Song, Yuexin Ma, Zhe Wang, Hongsheng Li, and Dahua
  Lin.
\newblock Cylinder3d: An effective 3d framework for driving-scene lidar
  semantic segmentation.
\newblock {\em arXiv preprint arXiv:2008.01550}, 2020.

\bibitem{Zhou_2021_CVPR}
Qiang Zhou, Chaohui Yu, Zhibin Wang, Qi~Qian, and Hao Li.
\newblock Instant-teaching: An end-to-end semi-supervised object detection
  framework.
\newblock In {\em Proc. of the IEEE/CVF Conference on Computer Vision and
  Pattern Recognition (CVPR)}, pages 4081--4090, June 2021.

\bibitem{multiviewfusion}
Yin Zhou, Pei Sun, Yu~Zhang, Dragomir Anguelov, Jiyang Gao, Tom Ouyang, James
  Guo, Jiquan Ngiam, and Vijay Vasudevan.
\newblock End-to-end multi-view fusion for 3d object detection in lidar point
  clouds.
\newblock In {\em CoRL 2019}, 2019.

\bibitem{zhu2021cylindrical}
Xinge Zhu, Hui Zhou, Tai Wang, Fangzhou Hong, Yuexin Ma, Wei Li, Hongsheng Li,
  and Dahua Lin.
\newblock Cylindrical and asymmetrical 3d convolution networks for lidar
  segmentation.
\newblock In {\em Proc. of the IEEE/CVF conference on computer vision and
  pattern recognition}, pages 9939--9948, 2021.

\bibitem{(AF)2-S3Net},
Cheng, R. and Razani, R. and Taghavi, E. and Li, E. and Liu, B.
\newblock (AF)2-S3Net: Attentive Feature Fusion With Adaptive Feature Selection for Sparse Semantic Segmentation Network.
\newblock In CVPR, 2021.

\bibitem{point-to-Voxel}
Hou, Y. and Zhu, X. and Ma, Y. and Loy, C. C. and Li, Y.
\newblock Point-to-Voxel Knowledge Distillation for LiDAR Semantic Segmentation
\newblock In CVPR, 2022.

\bibitem{nuScenes_2020}
Caesar, H. and Bankiti, V. and Lang, A. H. and Vora, S. and Liong, V. E. and Xu, Q. and Krishnan, A. and Pan, Y. and Baldan, G. and Beijbom, O.
\newblock nuScenes: A Multimodal Dataset for Autonomous Driving.
\newblock In CVPR, 2020.


\end{thebibliography}

%

\end{document}